\documentclass[lettersize,journal]{IEEEtran}
\usepackage{amsmath,amsfonts}
\usepackage{array}
\usepackage[caption=false,font=normalsize,labelfont=sf,textfont=sf]{subfig}
\usepackage{textcomp}
\usepackage{stfloats}
\usepackage{makecell}
\usepackage{url}
\usepackage{float}
\usepackage{verbatim}
\usepackage{cite}
\usepackage{graphicx}       
\usepackage{algorithm}      
\usepackage{placeins}       
\usepackage{amsmath}
\usepackage{algpseudocode}
\usepackage{arydshln}
\usepackage[absolute,overlay]{textpos}
\usepackage{booktabs}
\usepackage{pifont}  
\begin{document}

\begin{textblock*}{20cm}(1.37cm,0.94cm)
\fontsize{7}{6}\selectfont
This work has been submitted to the IEEE for possible publication. Copyright may be transferred without notice, after which this version may no longer be accessible.
\end{textblock*}

\title{FaceEditTalker: Controllable Talking Head Generation with Facial Attribute Editing}

\author{%
Guanwen Feng$^{1,2,3}$~\IEEEmembership{Student Member, IEEE}, 
Zhiyuan Ma$^{1,2,3}$, 
Yunan Li$^{1,2,3,*}$~\IEEEmembership{Member, IEEE}, 
Jiahao Yang$^{1,2}$, 
Junwei Jing$^{1,2,3}$, 
Qiguang Miao$^{1,2,3,*}$~\IEEEmembership{Senior Member, IEEE}%
\thanks{$^{1}$Xi’an Key Laboratory of Big Data and Intelligent Vision, Xidian University, Xi'an 710071, China.}%
\thanks{$^{2}$Key Laboratory of Collaborative Intelligence Systems, Ministry of Education, Xidian University, Xi'an 710071, China.}%
\thanks{$^{3}$School of Computer Science and Technology, Xidian University, Xi'an 710071, China.}%
\thanks{* Corresponding authors: Yunan Li (yunanli@xidian.edu.cn); Qiguang Miao (qgmiao@xidian.edu.cn)}%
}



\maketitle
\begin{abstract}
Recent advances in audio-driven talking head generation have achieved impressive results in lip synchronization and emotional expression. However, they largely overlook the crucial task of facial attribute editing. This capability is indispensable for achieving deep personalization and expanding the range of practical applications, including user-tailored digital avatars, engaging online education content, and brand-specific digital customer service. In these key domains, flexible adjustment of visual attributes, such as hairstyle, accessories, and subtle facial features, is essential for aligning with user preferences, reflecting diverse brand identities and adapting to varying contextual demands. In this paper, we present FaceEditTalker, a unified framework that enables controllable facial attribute manipulation while generating high-quality, audio-synchronized talking head videos. Our method consists of two key components: an image feature space editing module, which extracts semantic and detail features and allows flexible control over attributes like expression, hairstyle, and accessories; and an audio-driven video generation module, which fuses these edited features with audio-guided facial landmarks to drive a diffusion-based generator. This design ensures temporal coherence, visual fidelity, and identity preservation across frames. Extensive experiments on public datasets demonstrate that our method achieves comparable or superior performance to representative baseline methods in lip-sync accuracy, video quality, and attribute controllability. Project page: \url{https://peterfanfan.github.io/FaceEditTalker/}. We will release the source code to the public upon acceptance.
\end{abstract}


\section{Introduction}

\begin{figure*}
  \centering
  \includegraphics[width=0.9\textwidth]{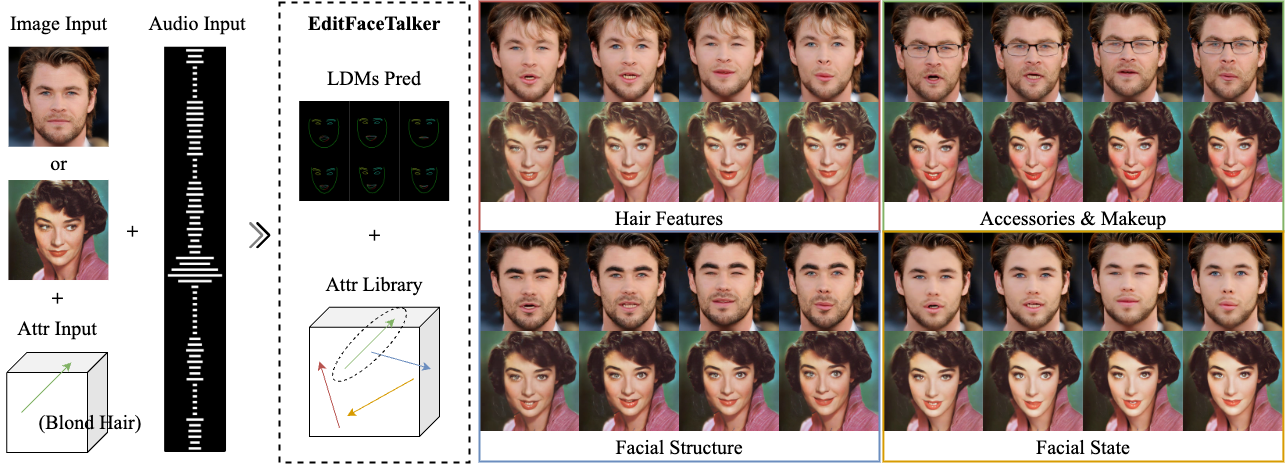}
  \caption{By providing a single reference image, audio input, and optional facial attribute input, our method generates high-quality, facially editable speaker videos by predicting facial landmark maps and performing linear edits on the feature semantic encoding of the image, combined with a diffusion model. This method demonstrates good generalization ability and achieves high lip-sync accuracy. In this figure, the image input used is a portrait from outside the dataset.}
\end{figure*}

\IEEEPARstart{I}{n} recent years, audio-driven talking head generation~\cite{Prajwal2020LipSync, Park2022SyncTalkFace, Cheng2022VideoRetalking, Guo2021ADNeRF, Shen2023SDNeRF, Li2023RegionAwareNeRF} has achieved remarkable progress and found widespread applications in domains such as virtual reality~\cite{li2024ae,jiang2024audio}, digital humans~\cite{Guo2021ADNeRF}, online education~\cite{jiang2024audio}, animation production~\cite{lan2023application}, and film post-production~\cite{li2024ae}. These methods enable virtual characters to synchronize facial movements with audio input, producing natural speaking behaviors. However, most existing approaches primarily focus on lip synchronization~\cite{Prajwal2020LipSync, Park2022SyncTalkFace, Cheng2022VideoRetalking, guan2023stylesync, yaman2024audiodriventalkingfacegeneration} and emotional expression~\cite{feng2024talker, wang2024emotivetalk, liang2024emotional, feng2024emospeaker,sheng2024Emotional,Lyu2025Emotional}, while largely overlooking the important functionality of controllable facial attribute editing.

Facial attribute editing is essential for audio-driven video generation due to its strong practical relevance. Beyond accurate audio-visual synchronization, users often require precise and flexible control over visual appearance, including expressions, hairstyles, age, gender, makeup, and accessories like glasses. For example, virtual idols may need to adapt to different audience preferences, and digital customer service agents may need to reflect their distinct brand identities. Dynamic and fine-grained attribute control can greatly enhance user engagement and personalization. 

Previous research on facial attribute editing has been extensive, initially focusing on static face images. GAN-based methods have achieved significant success in this domain, with representative examples including StyleGANs~\cite{Karras2018StyleGAN, Karras2019StyleGAN2}, which leverage a highly disentangled latent space to enable realistic and controllable facial edits. Naturally, researchers have attempted to extend these techniques to video generation, which introduces new challenges in maintaining facial detail, temporal consistency, and overall video quality. (1) \textbf{Poor facial detail:} Although GAN-based methods employ strategies such as frame alignment and fine-tuning to preserve temporal coherence during frame-by-frame editing, these approaches can still result in misalignment artifacts and inconsistent facial details~\cite{xu2022videoeditgan}, as well as background flicker~\cite{yin2022styleheat} and other visible video artifacts.
(2) \textbf{Temporal discontinuity in editing:} Existing methods often suffer from temporal artifacts, such as visual flickering~\cite{xu2022temporallyconsistentsemanticvideo} or fluctuations of dynamic attributes (e.g., beard, eyeglasses) during motion~\cite{yao2021latent,tzaban2022stitch}. Variations in head pose can further compromise temporal consistency, resulting in non-smooth or perceptually unstable edits. These limitations are largely attributable to the intrinsic capacity constraints of GAN-based models, which hinder their ability to accurately encode and transfer the complex information embedded in both source and target attribute frames. Although diffusion-based approaches generally yield higher video fidelity and more robust attribute manipulation than GAN-based methods, they are nevertheless susceptible to both imperfect facial detail~\cite{bounareli2025diffusionact} and temporal inconsistency~\cite{kim2023diffusion,li2025qffusion,bounareli2025diffusionact}.

\enlargethispage{-\baselineskip}

To address these limitations, we propose FaceEditTalker, a novel framework combining audio-driven talking head generation with controllable facial attribute editing. We adopt a dual-layer latent encoding structure~\cite{preechakul2022diffusion} to jointly model high-level semantics and low-level textures, where the semantic encoder conditions the reference image to guide DDIM for accurate facial reconstruction. A linear classifier is trained on the attribute semantic code to produce an attribute vector stored in a label-vector library; fusing this vector with the original semantic code yields an edited semantic code that guides DDIM to generate the desired attributes. Unlike StyleGAN-based methods~\cite{xu2022videoeditgan,tzaban2022stitch}, which often sacrifice reconstruction quality for precise editing, our approach achieves near-perfect face reconstruction while preserving fine-grained details. Furthermore, we adopt the landmark predictor from the representative method~\cite{wei2024aniportrait} to accurately infer landmark features, enabling joint guidance by semantic code and landmarks to prevent facial jitter and attribute fluctuations, ensuring temporal consistency. Our framework seamlessly integrates these components to achieve high-fidelity and editable talking head generation. Extensive experiments on multiple public datasets demonstrate superior performance in video quality, keypoint alignment, and identity preservation.


Our main contributions are summarized as follows:

\begin{itemize}
    \item We propose FaceEditTalker, the first framework that seamlessly unifies facial attribute editing and audio-driven talking head generation, enabling fine-grained manipulation of attributes such as hair, facial structure, and accessories, while maintaining natural lip movements and facial dynamics.
    \item We introduce a novel two-stage heterogeneous latent diffusion model to address the challenges of editing capability and consistency, enabling highly flexible zero-shot editing while effectively preserving identity integrity and temporal coherence.
    \item We conduct extensive evaluations on multiple public datasets, demonstrating that our method outperforms existing baselines in video quality, lip synchronization, keypoint alignment, and identity preservation.
\end{itemize}

\section{Related Work}




\subsection{Audio-driven Talking Head Generation.}  
Recent methods for audio-driven talking head generation have made remarkable progress, emphasizing realism, identity preservation, and expression diversity. Early approaches~\cite{Prajwal2020LipSync, Park2022SyncTalkFace, Cheng2022VideoRetalking} primarily adopt encoder-decoder architectures to map audio signals to lip movements. Although effective to some extent, these methods often suffer from blurred textures and weak identity preservation due to limited fusion strategies. To enhance realism, NeRF-based methods~\cite{Guo2021ADNeRF, Shen2023SDNeRF, Li2023RegionAwareNeRF, li2023efficient,li2023aenerfaudioenhancedneural} and 3D Gaussian Splatting methods~\cite{agarwal2025gensync, ye2024mimictalk, li2024talkinggaussian, feng2025gaussian,gong2025monocular,li2025instag} model 3D geometry for more lifelike appearances; however, they typically require long video sequences and come with high computational costs, limiting their practicality in real-time scenarios. Another line of work leverages facial landmarks or 3D priors~\cite{Chen2019HierarchicalTalkingFace, Zhou2020MakeItTalk,  Zhang2021FlowGuidedTalkingFace, Zhang2023SadTalker,chen2025echomimic} to disentangle speech content from identity features, improving controllability but often sacrificing fine-grained details in critical regions such as lips and teeth.
More recently, diffusion-based models~\cite{agarwal2025gensync,Shen2023DiffTalk,Tian2024EMO,Xu2024HALLO,he2024syncdiff,shen2025long,qiu2025skyreels,chatziagapi2025av,xu2025hunyuanportrait,hong2025audio,meng2025echomimicv2,meng2025echomimicv3,cui2024hallo2,hallo3} have emerged as a promising direction for high-quality and expressive talking head generation. Unlike GAN-based approaches that generate frames in a single forward pass, diffusion models iteratively denoise random noise under the guidance of conditioning signals such as audio and landmarks, enabling more precise control over motion dynamics and temporal consistency. Operating in structured control pipelines, these models are capable of synthesizing realistic lip movements, nuanced expressions, and coherent head pose.

In our method, facial landmarks are adopted as controllable priors to guide a diffusion-based generator, enabling precise lip synchronization and identity preservation while supporting flexible facial attribute editing without compromising speech-driven facial dynamics. Furthermore, the semantic code encoded from the reference image is incorporated as an additional control condition to guide the generation process, ensuring the preservation of fine-grained facial details. 

\begin{figure*}[h]
  \centering
  \includegraphics[width=0.9\textwidth]{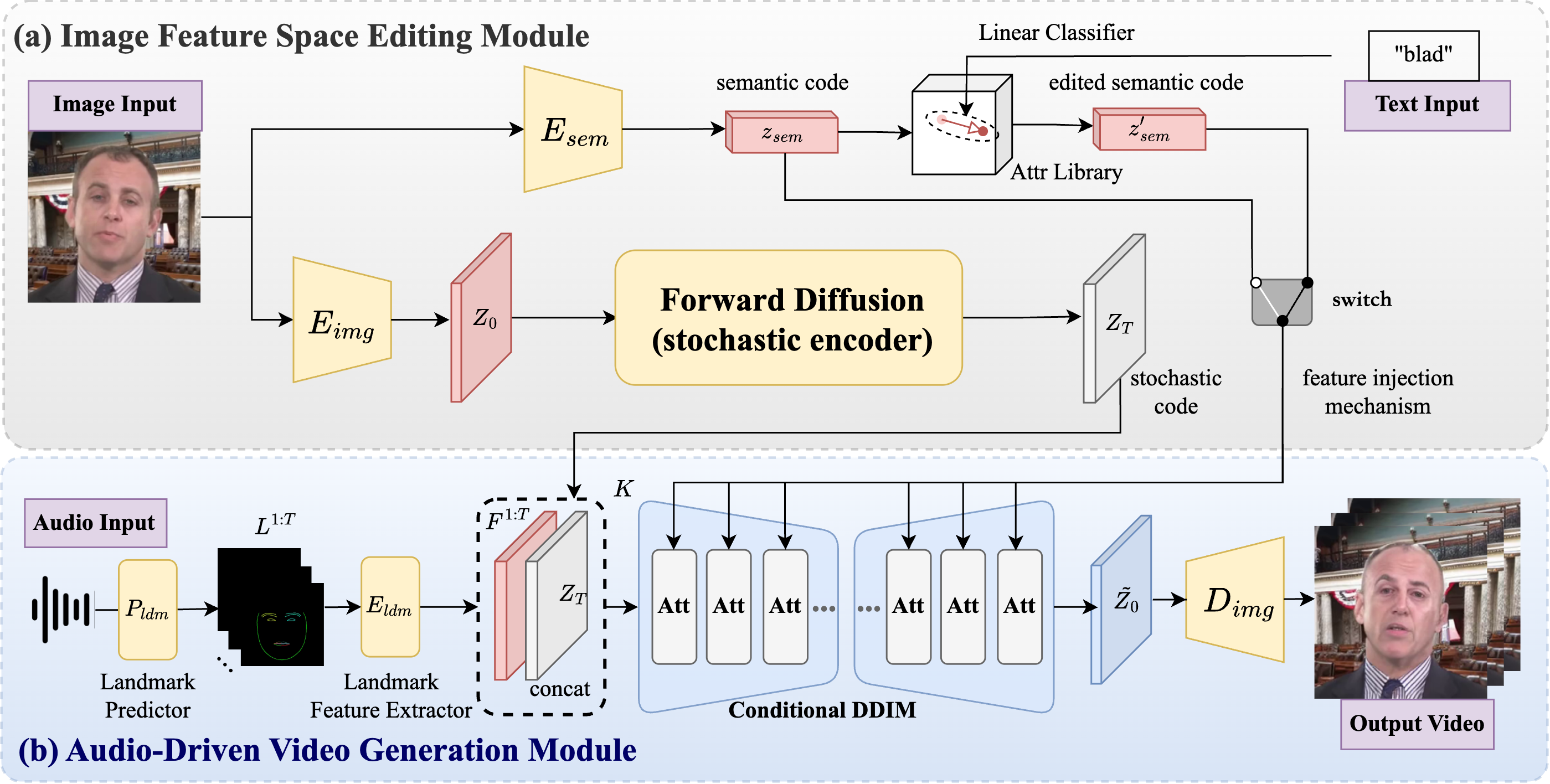}
  \caption{\textbf{Overview of the inference process of our proposed framework FaceEditTalker.} The framework consists of two main modules: \textbf{(a) Image Feature Space Editing Module}, which extracts editable semantic and stochastic codes from the reference image using a dual-layer latent encoding structure. Fine-grained attribute manipulation is enabled through optional spatial editing on the semantic codes. \textbf{(b) Audio-Driven Video Generation Module}, which leverages the audio input to infer driving landmarks. During the diffusion process, the stochastic codes guide dynamic generation, while the semantic codes serve as conditional inputs to ensure attribute consistency and visual fidelity throughout the video. 
The training procedure is detailed in Section \ref{sec:trainstratety} and Section \ref{sec:training-stage}.
}
\label{fig:Overview}
\end{figure*}

\subsection{Facial Attribute Editing.} 
Facial attribute editing focuses on modifying specific facial characteristics, such as age, hairstyle, and glasses, while preserving the subject’s identity. StyleGAN-based methods~\cite{Karras2018StyleGAN, Karras2019StyleGAN2, Abdal2019Image2StyleGAN, Richardson2020EncodingStyle} achieve controllable editing through latent space disentanglement, while CLIP-guided approaches~\cite{Radford2021CLIP, Patashnik2021StyleCLIP} introduce semantic alignment between text and images, enabling intuitive language-driven modifications. The emergence of diffusion models has further expanded the possibilities for producing realistic and expressive facial animations, offering enhanced editing control and fewer visual artifacts~\cite{preechakul2022diffusion, Banerjee2023Aging, Ding2023DiffusionRig}; however, extending these models to video sequences introduces challenges in maintaining temporal consistency due to frame-wise stochasticity, which can cause attribute variations across frames and lead to inconsistency. To address this, Latent Transformer~\cite{yao2021latent} performs optical flow alignment and refines latent codes in StyleGAN \(\mathcal{W}^+\) with identity- and attribute-preserving regularization to achieve temporally stable frame-wise editing. STIT~\cite{tzaban2022stitch} encodes adjacent frames into smoothly varying latent codes, enforces global identity via PTI, and performs stitching-based refinement to integrate edits without relying on explicit temporal loss. Diffusion Video Autoencoders~\cite{kim2023diffusion} represent a video using a shared identity vector and per-frame motion/background vectors; editing the shared identity ensures temporal coherence and consistent facial attributes across frames.

Despite these advances, achieving temporally consistent and high-fidelity facial attribute editing remains a key challenge in video-based editing. To address this, we introduce a motion module within the DDIM framework that employs Temporal Self-Attention across multiple resolutions with temporal position encoding to capture frame-to-frame dependencies. Attribute editing is performed in two stages: attribute vectors are first derived using a trained linear classifier, then used to adjust the semantic code toward the target attribute. The modified semantic code subsequently guides the DDIM to generate videos that retain the first-frame identity while exhibiting the desired attribute. This design enables precise, high-fidelity edits with coherent facial dynamics across frames.

\section{Method}
In this section, we provide a comprehensive and detailed description of the FaceEditTalker framework. Section~\ref{sec:Overview} presents an overview of the overall architecture. Section~\ref{sec:Task Formulation} formulates the task and outlines the workflow of the method. Section~\ref{sec:Image Feature Space Editing Module} details the image feature space editing module for facial attribute manipulation, while Section~\ref{sec:Audio-Driven Video Generation Module} describes the audio-driven video generation module. Finally, Section~\ref{sec:trainstratety} clarifies the training objectives, loss functions, and the inference procedure.
\label{sec:method}

\begin{figure*}[h]
  \centering
  \includegraphics[width=0.9\textwidth]{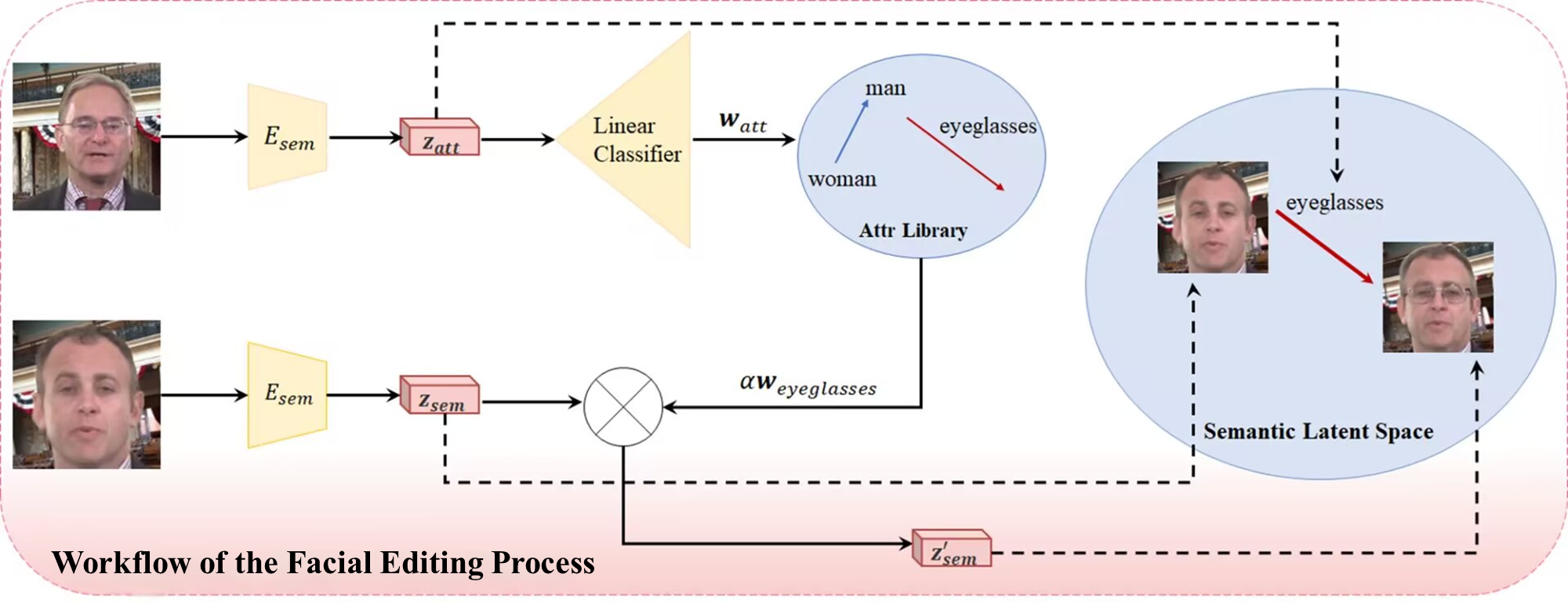}
  \caption{\textbf{Workflow of the Facial Editing Process.} The attribute direction vector $w_{att}$ is learned via a linear classifier. The original semantic code $z_{sem}$ is then linearly transformed along this direction with a strength factor $\alpha$ to produce the edited semantic code containing the desired attribute.}

\label{fig:Image Editing}
\end{figure*}

\subsection{Overview}
\label{sec:Overview}
Our proposed framework, FaceEditTalker, consists of two tightly coupled modules: the Image Feature Space Editing Module and the Audio-Driven Video Generation Module as shown in Fig.~\ref{fig:Overview}. The Image Feature Space Editing Module extracts editable semantic and stochastic codes from the reference image using a dual-layer latent encoding structure, enabling fine-grained control over facial attributes, which can be further edited using text-guided linear classifiers. These features are then passed to the Audio-Driven Video Generation Module, where synchronized audio-driven landmarks guide a diffusion-based generative process to produce high-quality, temporally coherent talking head videos with consistent identity and natural lip-sync.

\subsection{Task Formulation}
\label{sec:Task Formulation}

We first introduce the notations used in our formulation. Let $x_{ref} \in \mathbb{R}^{H \times W \times 3}$ denote the reference image of the target person, $A^{1:T} = (a^1, \dots, a^T)$ the extracted audio features, $y$ the facial attribute labels, $z_{\text{att}}$ the attribute code and $z_{\text{sem}}$ the semantic code. The driving landmark sequence is represented as $L^{1:T} = (l^1, l^2, \dots, l^T) \in \mathbb{R}^{T \times H \times W \times 3}$, and the generated video as $\hat{V} = \{ \hat{x}_1, \hat{x}_2, \dots, \hat{x}_T \} \in \mathbb{R}^{T \times H \times W \times 3}$.

For \textbf{facial attribute editing}, we leverage a labeled face dataset. Each image is encoded by a semantic encoder into an attribute vector $z_{\text{att}}$, which, together with its label $y$, is used to train a linear classifier $C$ that learns the mapping between labels and attribute vectors. During editing, $y$ serves as a key to retrieve the corresponding vector, and the semantic representation $z_{\text{sem}}$ is modified with a manipulation strength $\alpha$, formally expressed as
\begin{equation}
z_{\text{sem}}' = C(z_{\text{sem}}, y, \alpha).
\end{equation}

For \textbf{audio-driven talking head generation}, we adopt the pre-trained wav2vec model~\cite{baevski2020wav2vec} to extract audio features $A^{1:T}$, which are processed by a multiscale landmark prediction network to generate the driving landmark sequence $L^{1:T}$. Given $x_{ref}$ and $L^{1:T}$, the objective is to synthesize a realistic talking-head video $\hat{V}$ that preserves the identity of $x_{ref}$ while following the motion dynamics of $L^{1:T}$. This process is formulated as
\begin{equation}
\hat{V} = g(x_{ref}, L^{1:T}, z_{\text{sem}}'),
\end{equation}
where $g$ denotes the proposed generative model. Details of each component are presented in the following sections.


\subsection{Image Feature Space Editing Module}  
\label{sec:Image Feature Space Editing Module}
To achieve effective facial attribute editing, the Image Feature Space Editing Module leverages the design of DiffAE \cite{preechakul2022diffusion} with a dual-layer latent encoding structure. Inspired by the style vector mechanism in StyleGAN \cite{Karras2018StyleGAN}, our model decouples the latent space into two subspaces: semantic code \(z_{\text{sem}}\) and stochastic code \(Z_T\), capturing high-level semantic features and fine-grained details, respectively. This decomposition improves facial reconstruction accuracy and enhances controllability in attribute manipulation, supporting both zero-shot editing and fine-grained semantic control.

The \textbf{semantic encoder} \( E_{\text{sem}} \) extracts global facial semantics from the input image, encoding them into low-dimensional vectors akin to StyleGAN's style vectors, enabling linear transformations for attribute editing. As shown in Fig.~\ref{fig:Image Editing}, to enable text-driven editing and better align with the illustrated process, we employ the semantic encoder to extract semantic codes from images annotated with target attributes, thus constructing an attribute library. In the next step, the text input is processed in conjunction with the attribute library by a linear classifier to obtain the attribute direction vector $w_{\text{att}}$. The linear classifier is implemented as a single-layer MLP, and the attribute direction vector corresponds to its weight parameters. Subsequently, the original semantic code $z_{\text{sem}}$ is transformed into the dimension as $w_{\text{att}}$, enabling $w_{\text{att}}$ to perform a linear transformation that steers the semantic code towards the desired attribute. Given a semantic code \( z_{\text{sem}} \) and an attribute direction vector \( w_{\text{att}} \), attribute manipulation is expressed as:

\begin{equation}
z'_{\text{sem}} = z_{\text{sem}} + \alpha \cdot w_{\text{att}},
\end{equation}
where $\alpha$ controls the intensity of the change, and $z'_{\text{sem}}$ denotes the edited semantic code. 

In this attribute editing process, any of the 40 predefined attributes can be flexibly selected. By specifying the desired attribute at the corresponding code position, the editing process described above can be applied to achieve attribute manipulation on the image. Furthermore, it is also possible to design automated scripts to systematically generate edited results for all attributes. To ensure consistency in non-target areas, we use the reconstruction loss $L_{\text{rec}}$ to measure the difference between the edited latent code $z'_{\text{sem}}$ and the static code $z_{\text{static}}$ representing unedited regions. The reconstruction loss $L_{\text{rec}}$ is defined as:

\begin{equation}
L_{\text{rec}} = |z'_{\text{sem}} - z_{\text{static}}|.
\end{equation}


The \textbf{stochastic encoder}, designed as a Unet-based diffusion model \cite{song2020denoising}. First, the image encoder \(E_{img}\) encode the input image into an initial latent representation \(Z_0\). This representation undergoes a noise diffusion process to capture fine details, resulting in the final latent representation \(Z_T\). The parameter $\alpha_t$ controls the proportion of the original signal retained in each diffusion step, regulating the balance between signal preservation and noise injection during the forward process. We refer to this complete encoding and diffusion procedure as a stochastic encoder. The forward process is defined as:

\begin{equation}
z_{t+1} = \sqrt{\alpha_{t+1}} z_t + \sqrt{1 - \alpha_{t+1}} \epsilon, \quad \epsilon \sim \mathcal{N}(0, \mathbf{I})
\end{equation}

During the forward diffusion process, the initial latent representation $Z_0$, obtained from the input image, is progressively noised using our defined stochastic encoder. This encoder is specifically designed to implement the forward diffusion process, in which Gaussian noise is systematically added to $Z_0$ over a sequence of time steps according to a predefined noise schedule. As a result, the representation gradually transitions into a highly stochastic latent code $Z_T$, which captures rich local detail variations essential for realistic facial generation in the subsequent reverse process.

\subsection{Audio-Driven Video Generation Module}
\label{sec:Audio-Driven Video Generation Module}

As shown in Fig.~\ref{fig:Overview}, this module generates video using audio input $S_{\text{audio}}$, frame sequence $V = \{x_1, x_2, \dots, x_T\}$, semantic code $z_{\text{sem}}$ (irrespective of whether it is edited) and stochastic code $Z_T$. To perform audio-to-landmark prediction, we employ two networks: \textbf{a landmark predictor $P_{\text{ldm}}$ and a landmark feature extractor $E_{\text{ldm}}$.}

First, the audio input $S_{\text{audio}}$ is first processed by a pre-trained Wav2Vec model to extract audio features $A^{1:T}$: 

\begin{equation}
    A^{1:T}=Wav2Vec(S_{audio}).
\end{equation}

\textbf{Landmark Predictor:} Inspired by AniPortrait~\cite{wei2024aniportrait}, which demonstrates strong lip-sync accuracy and temporal consistency, we define an audio-driven landmark predictor $P_{\text{ldm}}$ to generate a facial landmark sequence $L^{1:T}$ from the audio features $A^{1:T}$ and the reference image $x_{\text{ref}}$:

\begin{equation}
    L^{1:T}= P_{\text{ldm}}(A^{1:T},x_{\text{ref}}).
\end{equation}

\textbf{Landmark Feature Extractor:} The facial landmark sequence $L^{1:T}$ is processed by a landmark feature extractor to obtain the corresponding landmark features $F^{1:T}$:

\begin{equation}
F^{1:T} = E_{ldm}(L^{1:T},x_{ref}),
\end{equation}
where $E_{\text{ldm}}$ employs multiscale strategies and cross-attention mechanisms to fuse the landmark sequence $L^{1:T}$ with the reference landmark sequence $l_{\text{ref}}$ extracted from the reference image $x_{\text{ref}}$, producing landmark features $F^{1:T}$. This design enables the model to capture multilevel facial dynamics and strengthen feature correlations, thus improving precision and temporal consistency in video generation, as further validated by the ablation results reported in Table~\ref{tab:freq}.

Subsequently, the facial landmark features $F^{1:T}$ are fused with the stochastic code $Z_T$ through a residual connection, producing the feature representation $K$, which is then used as the input to the diffusion model:

\begin{equation}
K=Concat(F^{1:T},Z_T)
\end{equation}

During the diffusion model sampling process, we employ conditional DDIM~\cite{song2022ddim}, where high-level semantic information \(z_{\text{sem}}\) is incorporated as a global facial attribute control signal, while the feature \(K\) provides dynamic motion information. This design ensures stable expression of semantic attributes alongside synchronized audio-driven facial movements. The DDIM reverse sampling at timestep \(t\) is formulated as:

\begin{equation}
z_{t-1} = \sqrt{\alpha_t} z_t + \sqrt{1-\alpha_t} \epsilon_\theta(z_t, z_{\text{sem}}, F, t),
\end{equation}
where $z_t$ represents the noisy data in the timestep $t$, and $\epsilon_\theta$ denotes the conditional denoising network that guides the denoising process based on the latent variable \( z_{\text{sem}} \) and the input feature \( K \). The conditional feature injection in the diffusion model employs a dual mechanism:

\begin{itemize}
    \item \textbf{Semantic code \( z_{\text{sem}} \)}: injected via cross-attention layers to provide global facial attribute control.
    \item \textbf{Input feature \( K \)}: fused from the inputs to deliver local motion control.  
\end{itemize}

Furthermore, multiscale feature fusion ensures that control signals are effectively propagated across different resolution levels, enhancing both global semantic consistency and fine-grained motion synchronization.

Finally, the denoised latent variable $Z_0$ is decoded into a sequence of video frames that not only exhibit dynamic facial expressions synchronized with the input audio, but also enable controllable facial attribute editing based on the selected semantic encoding $z_{\text{sem}}$, depending on the switch configuration. This process produces high-quality editable talking head videos. For further implementation details, please refer to Section \ref{sec:inference-stage}.

\subsection{Training and Inference Pipeline}
\label{sec:trainstratety}
\textbf{Training process}: Our model employs a three-stage training framework to progressively learn semantic representation, attribute classification, and conditional generation. Each stage builds upon the previous to ensure robust feature learning, precise attribute control, high-quality and controllable synthesis.

\textbf{The first stage} serves as a pre-training phase, during which the semantic and stochastic encoders are jointly trained by minimizing the mean squared error between the predicted and ground-truth noise:
\begin{equation}
\mathcal{L}_{\text{sample}}=\sum_{t=1}^{T}\mathbb{E}_{x_{ref},\epsilon_{t}}[||\epsilon_{\theta}(z_{t},z_{\text{sem}},t)-\epsilon_{t}||_{2}^{2}],
\end{equation}
where \(\epsilon_\theta(\cdot)\) denotes the conditional denoising network that predicts the noise component, and \(\epsilon_{t}\) represents the ground-truth noise at timestep \(t\).

\textbf{The second stage} involves training a linear classifier using cross-entropy loss to accurately predict image attributes:
\begin{equation}
\mathcal{L} = -\frac{1}{N} \sum_{i=1}^{N} \sum_{c=1}^{C} y_{i,c} \log(\hat{y}_{i,c}),
\end{equation}
where \(N\) denotes the total number of samples, \(C\) is the number of attribute categories, and \(y_{i,c}\) and \(\hat{y}_{i,c}\) correspond to the ground truth labels (encoded as one-hot vectors) and the predicted attribute probabilities, respectively.

\textbf{The third stage} involves training a conditional diffusion model in the latent space. At each diffusion step, the model receives a noisy latent vector \(Z_T\) along with the semantic code \(z_{\text{sem}}\) and motion lanmark features \(F\). These conditioning signals are fused to guide the denoising network, allowing precise control over both static attributes and dynamic motion. This design facilitates stable training and enables high-quality, editable talking head synthesis. The training objective is to minimize the mean squared error between the predicted and ground-truth noise:
\begin{equation}
\mathcal{L}_{\text{sample}}=\sum_{t=1}^{T}\mathbb{E}_{x_{ref},\epsilon_{t}}[||\epsilon_{\theta}(z_t,z_{\text{sem}},F,t)-\epsilon_{t}||_{2}^{2}],
\end{equation}
where the semantic and motion features are fused to condition the latent diffusion process, enabling the model to jointly perform facial attribute editing and dynamic motion synthesis within a stable and efficient latent space.

\textbf{Inference process}: We generate talking head videos using only an input audio clip and a reference image, with optional attribute specifications for editing. During inference, the audio and reference image are used to extract the landmark feature $F$, while the reference image also provides a semantic code $z_{sem}$. If editing is required, the semantic code is fused with the corresponding attribute vector $w_{att}$. The landmark feature $F$ is then combined with the stochastic code $Z_T$ and fed into the DDIM sampler, which performs denoising under the guidance of the semantic code $z_{sem}$ on the latent embedding that encodes landmark information. Finally, this process yields a temporally consistent sequence of video frames, where facial movements are synchronized with the audio and facial attributes are controllably edited as specified, resulting in high-quality, realistic, and editable talking head videos.

\section{Pseudo-code for this method}
\label{appendix:Pseudo-code}
To help researchers better understand the overall workflow of our approach, we present the pseudo-code along with a description of the training and inference processes.
\subsection{Training and Inference Process Description}
The method is organized into three principal training modules. The first two stages focus on training the Image Feature Space Module, while the third stage trains the Audio-Driven Video Generation Module, followed by the inference process.

\begin{enumerate}
    \item \textbf{Joint Training of Semantic Encoder and Stochastic Encoder}: 
    Face images are encoded to extract high-level semantic and stochastic features, optimized using a diffusion model.

    \item \textbf{Training the Image Semantic Linear Classifier}: 
    Attribute variation directions are learned in the high-level semantic space, enabling precise attribute classification.

    \item \textbf{Training the Audio-Driven Video Generation Module}: 
    The injected noise, together with landmark features, is incorporated into DDIM, where the generation process is guided by semantic code to produce high-quality, identity-consistent talking face videos.

    \item \textbf{Inference Process}: 
    In the inference phase, the model uses the input audio sequence, reference image, and attribute information to generate facial motion features, which are processed by the diffusion model to produce high-quality editable talking face videos.
\end{enumerate}

\subsection{Training Stage}
\label{sec:training-stage}

\begin{algorithm}[H]
\caption{Joint Training of Semantic Encoder and Stochastic Encoder}
\begin{algorithmic}[1]
\State \textbf{Input:} Face image set (with attribute labels), attribute list, learning rate, diffusion time step
\State \textbf{Output:} Weight vectors corresponding to each attribute list
\For{each epoch}
    \For{each \( x_i \in \text{FFHQ(image)} \)}
        \State $z_{\text{sem}} = \text{SemanticEncode}(x_i)$ \Comment{Semantic Encoder Forward Pass}
        \State $t = \text{RandomTimeStep}()$
        \State $Z_0 = \text{T=ImgEncoder}(x_i)$
        \State $Z_T = \text{AddNoise}(Z_0)$
        \State $\hat{x} = \hat{q}(Z_T, z_{\text{sem}})$ \Comment{Model Forward Pass}
        \State $e = \text{ComputeTarget}(Z_t)$ \Comment{Compute Target}
        \State $L = \text{MSE}(e, \hat{x})$ \Comment{Loss Calculation}
        \State \text{Backpropagate and update parameters}
    \EndFor
    \State \text{Save model at the end of each epoch}
\EndFor
\end{algorithmic}
\end{algorithm}

\begin{algorithm}[H]
\caption{Training the Image Attributes Linear Classifier}
\begin{algorithmic}[1]
\State \textbf{Input:} Face image set with facial attributes, attribute labels, learning rate
\State \textbf{Output:} Weight vectors corresponding to each attribute list
\For{each epoch}
    \For{each attribute \( x_i \in \text{Att}(x_1, x_2, \dots, x_n) \)}
        \State $z_{\text{att}} = \text{SemanticEncoder}(x_i)$
        \State $y_{\text{labels}} = \text{GetAttributeLabels}(x_i, \text{FFHQ}(image))$
        \State $W_{att} = \text{InitializeWeightVector}()$
        \State $b = \text{InitializeBias}()$
        \For{each $x_i \in \text{FFHQ(image)}$}
            \For{each $z_{\text{att}}$}
                \State $\hat{y_{i}} = \text{Sigmoid}(W_{att} z_{\text{att}} + b)$ \Comment{Forward Pass}
                \State $L = \text{CrossEntropyLoss}(\hat{y_i},y_i)$ \Comment{Cross-Entropy Loss}
                \State \text{Backpropagate and update parameters}
            \EndFor
        \EndFor
    \EndFor
    \State \text{Save weight vectors and attribute-label pairs $(y_i,w_{att,i})$}
\EndFor
\end{algorithmic}
\end{algorithm}
\begin{algorithm}[H]
\caption{Training the Audio-Driven Video Generation Module}
\begin{algorithmic}[1]
\State \textbf{Input:} reference image set, audio sequence $S_{audio}$, learning rate
\State \textbf{Output:} Diffusion model parameters
\State \textbf{Data Preprocessing:} Use Wav2Vec to extract audio feature $A^{1:T}$ sequence from audio sequence $\mathcal{S}_{audio}$
\For{each epoch}
    \For{each batch $A^{1:T},~x_{\text{ref}}$}
        \State $Z_0 = \text{ImgEncoder}(x_{ref})$
        \State $Z_T = \text{AddNoise}(Z_0)$
        \State $L^{1:T}=LandmarkPredictor(A^{1:T})$
        \State $F^{1:T}=LandmarkFeatureExtractor(L^{1:T})$
        \State $K=Concat(F^{1:T},Z_T)$
        \State $z_{\text{sem}} = \text{SemanticEncoder}(x_{\text{ref}})$
        \State $t = \text{RandomTimeStep}()$
        \State $\hat{\epsilon} = \epsilon_\theta(K, t \mid z_{\text{sem}})$
        \State $L = \text{MSE}(\hat{\epsilon}, \epsilon)$\Comment{MSE Loss}
        \State \text{Backpropagate and update parameters}
    \EndFor
    \State \text{Save model at the end of each epoch}
\EndFor
\end{algorithmic}
\end{algorithm}

\subsection{Inference Stage}
\label{sec:inference-stage}
\begin{algorithm}[H]
\caption{Controllable Talking Head Generation with Facial Attribute Editing}
\begin{algorithmic}[1]
\State \textbf{Input:} Reference image $x_{ref}$, audio sequence $\mathcal{S}_{audio}$, attribute label-weight pairs $(y_i, w_{\text{att,i}})$, attribute editing magnitude $\alpha$, diffusion time steps
\State \textbf{Output:} Speaker video frame sequence $S_{video}$

\State $A^{1:T}=A(a_{1},a_{2},...,a_{t})=Wav2Vec(S_{audio})$
\State $L^{1:T}=LandmarkPredictor(A^{1:T})$
\State $F^{1:T}=LandmarkFeatureExtractor(L^{1:T})$
\State $K=Concat(F^{1:T},Z_T)$
\State $z_{sem}=SemanticEncoder(x_{ref})$
\If{$\alpha$ is not None}
\State $z_{sem}=z_{sem}+\alpha \cdot w_{att}$
\EndIf
\State $S_{video}=DiffusionModel(K, z_{sem}, steps)$
\end{algorithmic}
\end{algorithm}


\section{Experiments}
\label{sec:Experiment}
\subsection{Experimental Settings}

\textbf{Data Preprocessing.} During training and validation, videos were sampled at 25 frames per second (FPS) and audio at 16 kHz. To ensure data consistency, videos were cropped and resized to a resolution of $512 \times 512$ pixels. Audio–video synchronization was achieved using Mel-spectrogram representations with a window length and hop length of 640 samples. In stage one, we used 16 input frames with a stride of 4; in stage three, 16 input frames with a stride of 1 and stride augmentation were employed.

\textbf{Training Configuration.} Our method was trained on two NVIDIA A100 GPUs. The first and second stages were trained for 100 hours and the third stage was trained for 160 hours. The main parameters are shown in Table~\ref{tab:main_parameters}:

\begin{table}[H]
\centering
\caption{Experimental Parameter Settings.}
\label{tab:main_parameters}
\begin{tabular}{l|l}
\hline
\textbf{Parameters} & \textbf{Value/Range} \\
\hline
Random Seed & 0 \\
Image Size & 512*512 \\
Batch Size & 16 \\
Learning Rate & 0.0001 \\
Training Epochs & 20000 \\
Embedding Layer Channels & 512 \\
Diffusion Timesteps & 1000 \\
\hline
\end{tabular}
\label{tab:experimental_parameters}
\end{table}

\textbf{Datasets.} We trained the dual-layer latent architecture encoder using the FFHQ dataset \cite{Karras2019StyleGAN2}, which offers high-resolution facial images with various attributes including age, race, expression, facial structure, hair features and accessories, ideal for learning complex feature representations. For the linear classifier, we used the CelebA-HQ dataset \cite{lee2020maskgan} with binary labels for 40 facial attributes to enhance attribute feature separation and model generalization. In the audio-driven facial animation generation stage, we utilized the HDTF dataset \cite{zhang2021flow}, containing lip-sync videos from more than 300 speakers, along with VoxCeleb2 \cite{nagrani2020voxceleb} and VFHQ \cite{xie2022vfhq} datasets to improve the model's ability to learn complex mappings between speech and facial movements under various environmental conditions. Additionally, we applied LatentSync \cite{li2024latentsync} to refine dataset quality by resampling videos, removing those with low synchronization confidence, correcting audiovisual offsets, and filtering out clips with poor HyperIQA scores, thereby enhancing lip-sync accuracy and visual quality.

\begin{table*}[h]
\centering
\caption{Quantitative evaluation of our approach compared with representative approaches.}
\resizebox{\textwidth}{!}{
\label{tab:quantitative evaluation}
\begin{tabular}{l c c c c c c c c c}
\toprule
\multicolumn{1}{c}{} & \multicolumn{4}{c}{Video Quality} & \multicolumn{3}{c}{Lip-sync} & \multicolumn{2}{c}{Keypoint Error} \\
\cmidrule(lr){2-5} \cmidrule(lr){6-8} \cmidrule(lr){9-10}
Method & FID $\downarrow$ & SSIM $\uparrow$ & PSNR $\uparrow$ & CPBD $\uparrow$ & Min Dist $\downarrow$ & AVConf $\uparrow$ & AVOffset($\rightarrow 0$) & M-LMD $\downarrow$ & F-LMD $\downarrow$ \\
\midrule
Real Video (HDTF) & 0.000 & 1.000 & 35.668 & 0.263 & 7.238 & 8.993 & 0.000 & 0 & 0 \\
\midrule
Wav2Lip\cite{Prajwal2020LipSync} & 20.641 & 0.532 & 16.929 & 0.199 & \textbf{6.611} & \textbf{8.119} & -2.000 & 4.368 & 4.256 \\
SadTalker\cite{Zhang2023SadTalker} & 25.566 & 0.698 & 22.211 & 0.204 & 8.527 & 3.163 & 1.000 & 3.368 & 3.192 \\
DiffTalk\cite{Shen2023DiffTalk} & 18.570 & 0.558 & \textbf{26.587} & \textbf{0.225} & 10.091 & 3.046 & -4.000 & 5.473 & \textbf{1.146} \\
EchoMimic\cite{chen2025echomimic} & 17.486 & \textbf{0.893} & 25.968 & 0.210 & 9.163 & 6.146 & -1.000 & 3.983 & 3.790 \\
Hallo\cite{Xu2024HALLO} & 16.880 & 0.821 & 25.331 & 0.203 & 9.612 & 6.128 & \textbf{0.000} & 3.412 & 3.532 \\
Our Method & \textbf{16.580} & 0.843 & 25.574 & 0.205 & 9.527 & 6.354 & \textbf{0.000} & \textbf{3.354} & 3.465 \\
\midrule
Real Video (VoxCeleb2) & 0.000 & 1.000 & 26.453 & 0.272 & 7.701 & 6.365 & 0.000 & 0 & 0 \\
\midrule
Wav2Lip\cite{Prajwal2020LipSync} & 20.565 & 0.468 & 16.042 & 0.201 & \textbf{7.665} & \textbf{8.236} & -2.000 & 4.368 & 4.256 \\
SadTalker\cite{Zhang2023SadTalker} & 23.421 & 0.634 & 21.254 & \textbf{0.211} & 13.542 & 3.355 & 1.000 & 3.368 & \textbf{3.192} \\
EchoMimic\cite{chen2025echomimic} & 17.586 & \textbf{0.910} & 24.948 & 0.209 & 9.654 & 6.542 & -1.000 & 3.983 & 3.790 \\
Hallo\cite{Xu2024HALLO} & 15.785 & 0.751 & 25.738 & 0.188 & 8.142 & 6.105 & \textbf{0.000} & 3.408 & 3.498 \\
Our Method & \textbf{15.418} & 0.772 & \textbf{25.985} & 0.189 & 8.068 & 6.252 & \textbf{0.000} & \textbf{3.354} & 3.465 \\
\bottomrule
\end{tabular}
}
\end{table*}

\textbf{Comparison Methods.} To the best of our knowledge, there is no existing method capable of generating high-resolution, audio-driven speaker videos with editable facial attributes. For a comprehensive evaluation of our proposed method, we first generate results using semantic features extracted by the high-level semantic encoding module, ensuring identity consistency with reference images. We compare our method with several representative and widely used lip synchronization methods. Wav2Lip \cite{Prajwal2020LipSync} optimizes direct mappings between audio and lip motion for highly synchronized lip movements while preserving facial textures. SadTalker \cite{Zhang2023SadTalker} employs explicit facial landmarks and adversarial networks to produce smooth animations. DiffTalk \cite{Shen2023DiffTalk}, EchoMimic \cite{chen2025echomimic}, and Hallo \cite{Xu2024HALLO} leverage diffusion models to model conditional distributions between audio and facial movements, achieving higher-quality talking videos and strong generalization capabilities for out-of-distribution subjects. This comparison aims to evaluate our method's performance relative to current leading techniques in audio-driven talking head generation.

\textbf{Evaluation Metrics.} For evaluating our method, we employ several metrics. Image generation quality is assessed using FID \cite{dowson1982frechet}, SSIM \cite{assessment2004error}, PSNR \cite{jahne2005digital}, and CPBD \cite{narvekar2011no}. Lip motion accuracy is evaluated with M-LMD and F-LMD \cite{chen2018lip}, while Syncconf \cite{prajwal2020lip} measures lip movement-audio synchronization. Additionally, we edit semantic features from the dual-layer semantic encoding module using a linear classifier to produce edited video results. These are compared against representative video editing methods such as Latent-Transformer \cite{yao2021latent}, STIT \cite{tzaban2022stitch}, and Diffusion-Video-Autoencoders \cite{kim2023diffusion}, using TL-ID and TG-ID \cite{tzaban2022stitch} as evaluation metrics.

\subsection{Performance Comparison}
\begin{figure*}[h]
  \centering
  \includegraphics[width=0.93\textwidth]{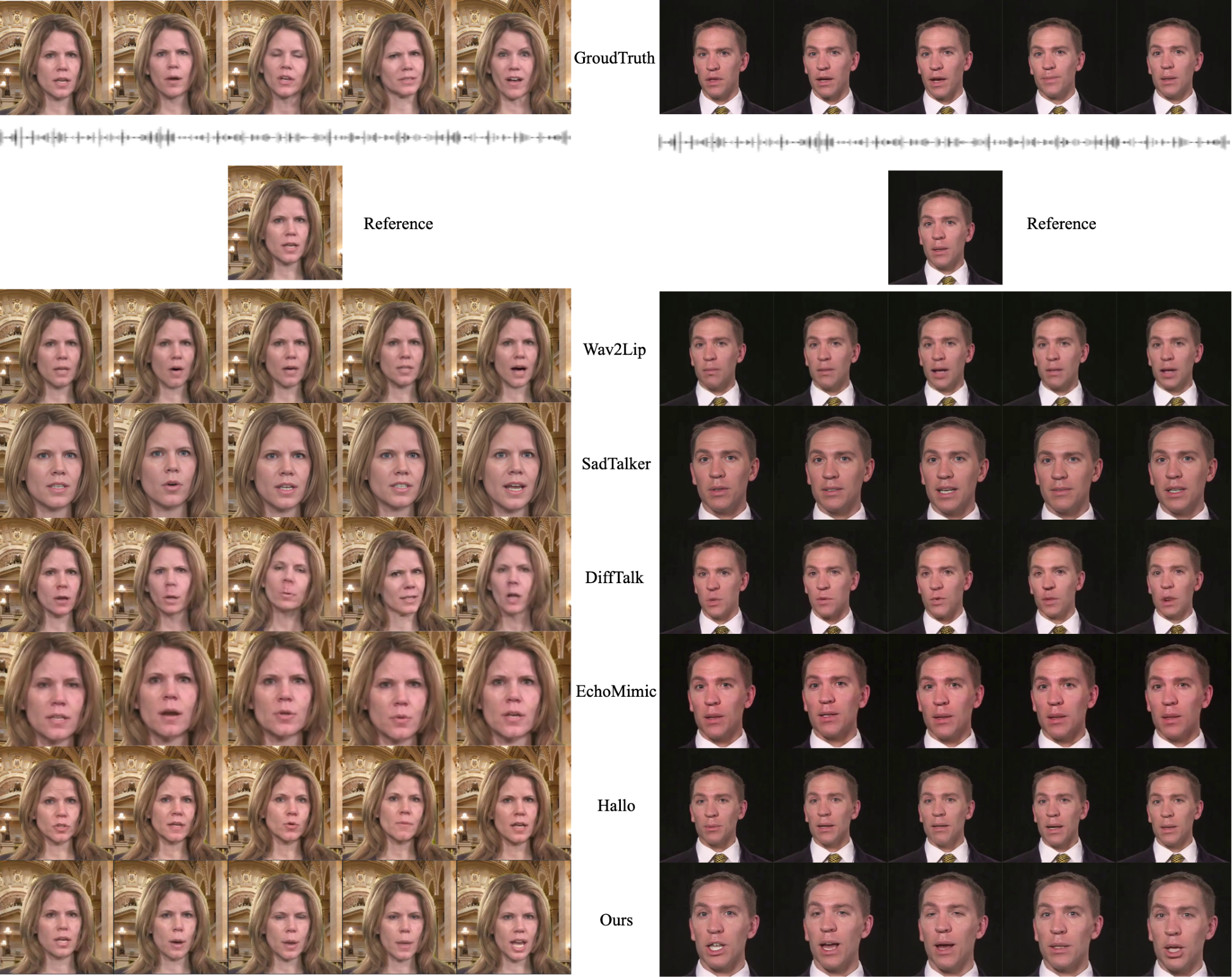}
  \caption{\textbf{Qualitative evaluation compared with other methods.} Using two different reference images and the same audio clip, our method is tested without enabling the editing feature. Our approach demonstrates superior performance in both facial expression naturalness and video quality.}
  \label{fig:qualitative}
\end{figure*}

\textbf{Quantitative Evaluation.} We quantitatively compared FaceEditTalker with representative audio-to-face generation methods on the HDTF and VoxCeleb2 datasets. As shown in Table~\ref{tab:quantitative evaluation}, FaceEditTalker achieves the best FID and M-LMD scores on both datasets, while maintaining competitive SSIM (second only to EchoMimic), PSNR, and lip-sync performance. This demonstrates strong overall performance across image feature similarity, structural similarity, and visual quality, enabled by the latent space diffusion model for fine-grained facial attribute control. Compared to prior diffusion-based approaches, our method benefits from a more advanced framework with enhanced semantic editing capabilities. Superior visual quality and robust audio-visual alignment, reflected in high SyncNet scores, are further supported by dataset corrections to reduce alignment errors, multi-scale modeling of global and local facial motions, and cross-attention mechanisms reinforcing audio–facial correspondence.

Despite these strengths, our method achieves moderate performance on metrics evaluating local lip articulation and fine-grained audio-visual alignment. Specifically, Min Dist indicates limited precision in modeling subtle lip movements, AVConf reflects suboptimal audio-lip synchronization, and F-LMD suggests reduced fidelity in reconstructing local lip contours. These limitations likely arise from the landmark-guided diffusion framework emphasizing global facial consistency, which can smooth localized lip variations, and the lack of fine-grained supervision in mapping audio features to lip shapes. Future work incorporating localized lip refinement, enhanced audio-lip feature mapping, and fine-grained supervisory signals could further improve lip motion fidelity and overall audio-visual coherence.


\textbf{Identity Consistency Evaluation.} We quantitatively evaluated identity consistency in face attribute editable talking head generation, as presented in Table~\ref{tab:identity and consistency evaluation}.  Our method edits semantic features to generate videos with 20 attributes, compared to video editing algorithms. Evaluation of identity consistency between frames showed that while our generative model achieved similar overall identity consistency as video editing methods, it significantly excelled in TL-ID.

\begin{table}[h]
\centering
\caption{Quantitative results of our approach compared with representative approaches. }
\label{tab:identity and consistency evaluation}
\begin{tabular}{l c c}
\toprule
Method & TL-ID $\uparrow$ & TG-ID $\uparrow$ \\
\midrule
Latent-Transformer\cite{yao2021latent} & 0.975 & 0.913 \\
STIT\cite{tzaban2022stitch} & 0.990 & 0.969 \\
Diffusion-Video-Autoencoders\cite{kim2023diffusion} & 0.986 & \textbf{0.991} \\
Our Method & \textbf{0.992} & 0.989 \\
\bottomrule
\end{tabular}
\end{table}

\begin{figure*}[h]
  \centering
  \includegraphics[width=0.93\textwidth]{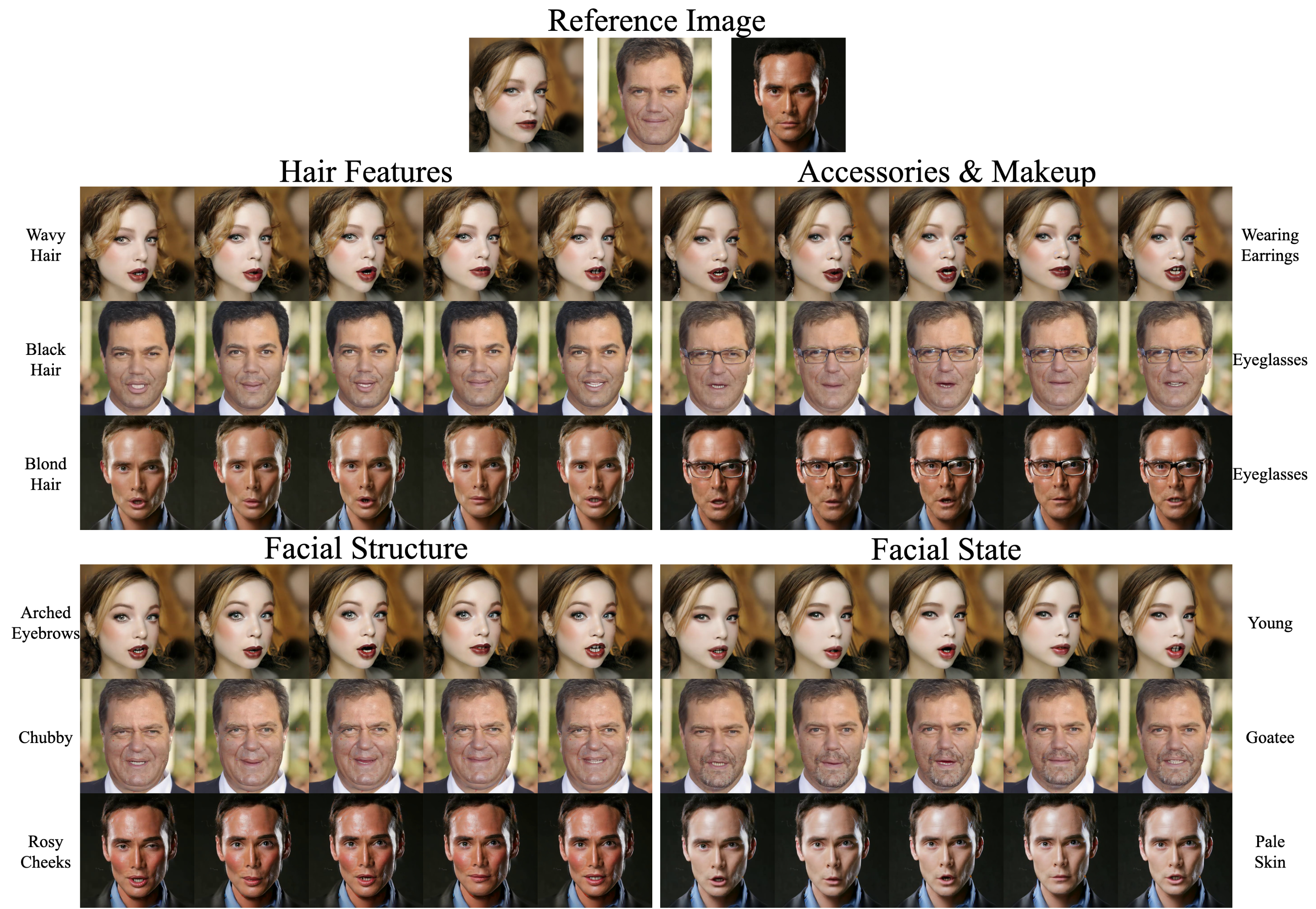}
  \caption{\textbf{Video generation results with the editing feature enabled.} Using three different reference images and the same audio clip, we demonstrate the editing and speaker generation effects under four different attribute editing categories with various sub-attributes.}
\end{figure*}


\textbf{Qualitative Evaluation.} Fig.~\ref{fig:qualitative} compares the generation quality of FaceEditTalker with existing advanced methods. While previous approaches often prioritize specific aspects such as lip synchronization or introduce artifacts and distortions when aiming for greater expressiveness, our proposed algorithm generates videos with better overall image quality and more accurately captures fine facial expression details, closely matching the original video. FaceEditTalker particularly excels at handling nuanced facial actions like eye closure and mouth opening, thereby contributing to a higher level of realism in the generated results.

\textbf{User Study.} To comprehensively evaluate the quality of the generated talking head videos, we conducted a user study involving 50 participants, focusing on four key dimensions: lip-sync accuracy, realism, video quality, and attribute editing effects. Each dimension was rated independently using a five point Likert scale (1 = poor, 5 = excellent). To ensure fairness and consistency, all methods—including our proposed approach and several baseline systems—were applied to the same source video under identical conditions. For methods supporting facial attribute editing, we introduced a predefined modification—the addition of eyeglasses—chosen for its visual saliency and broad applicability across different models. In contrast, methods lacking editing capability produced unaltered outputs. Participants were shown the generated videos and asked to assess each quality dimension separately. 

\setlength{\tabcolsep}{3pt}
\begin{table}[h]
\scriptsize
\centering
\caption{User Study Results for Editable Facial Attribute Talking Head Generation.}
\label{tab:user_study}
\begin{tabular}{l c c c c}
  \toprule
  Method & Lip-sync $\uparrow$ & Realism $\uparrow$ & Video Quality $\uparrow$ & \makecell{Attribute\\Editing Effect $\uparrow$} \\
  \midrule
  Original Video & 4.80 & 4.90 & 4.80 & \ding{55} \\
  Lip\cite{Prajwal2020LipSync} & \textbf{3.82} & \textbf{3.38} & 3.46 & \ding{55} \\
  SadTalker\cite{Zhang2023SadTalker} & 3.32 & 3.24 & 3.28 & \ding{55} \\
  DiffTalk\cite{Shen2023DiffTalk} & 2.96 & 3.30 & 2.96 & \ding{55} \\
  EchoMimic\cite{chen2025echomimic} & 3.48 & 3.34 & 3.06 & \ding{55} \\
  Hallo\cite{Xu2024HALLO} & 3.44 & 3.30 & 3.32 & \ding{55} \\
  Latent-Transformer\cite{yao2021latent} & 3.34 & 3.32 & 3.30 & 3.48 \\
  STIT\cite{tzaban2022stitch} & 3.08 & 3.18 & 2.96 & 3.04 \\
  \makecell[l]{Diffusion Video \\ Autoencoders\cite{kim2023diffusion}} & 2.96 & 3.22 & 3.44 & 3.26 \\
  Our Method & 3.58 & 3.18 & \textbf{3.50} & \textbf{4.04} \\
  \bottomrule
\end{tabular}
\end{table}

In the study, lip-sync accuracy was evaluated based on lip movement alignment with speech and consistency with the original video. Our method utilizes the Wav2Vec framework to extract audio features for predicting facial mesh and keypoints. Leveraging multi-scale keypoints and SyncNet preprocessing, it achieves strong lip-sync with a mean score of 3.58, second only to Wav2Lip. Realism and video quality were judged by the naturalness of expressions and visual clarity. The realism score averaged 3.18, limited by reliance on audio features alone, which sometimes caused rigid or less expressive facial movements. Video quality scored slightly higher at 3.50, benefiting from our latent diffusion architecture. For attribute editing, participants rated perceptual clarity and controllability; our method scored 4.04, outperforming most baselines—some of which lacked editing support (‘\ding{55}’ in Table~\ref{tab:user_study}). Overall, as shown in Table~\ref{tab:user_study}, our approach surpasses most baselines, demonstrating its effectiveness in generating realistic, high-quality, and controllably editable talking head videos.


\subsection{Analysis and Ablation Study}


\textbf{Effectiveness of the Facial Landmark Feature Extractor.}
We conducted ablation experiments on two datasets to validate the effectiveness of the proposed components in the facial landmark feature extractor. As illustrated in Table~\ref{tab:freq}, I denotes the Original Video serving as the ground-truth reference, II corresponds to the Multi-layer Convolution baseline, III represents the Multi-scale Strategy, IV indicates the Multi-scale combined with Cross-Attention, and V denotes the Edited Semantic Encoding. The results demonstrate that the multi-scale strategy (III) facilitates more accurate modeling of fine-grained facial dynamics such as lip and eye motions, while the integration of cross-attention (IV) further improves all evaluation metrics, particularly AVOffset and M-LMD, thereby enhancing temporal consistency with the original video. Moreover, the Edited Semantic Encoding (V) shows minimal impact on the quantitative evaluation metrics.

\begin{table}[h]
  \caption{Quantitative Metrics for Ablation Study of Facial Landmark Feature Extractor.}
  \resizebox{0.5\textwidth}{!}{
  \label{tab:freq}
  \begin{tabular}{l c c c c c}
    \toprule
    Method & Min Dist $\downarrow$ & AVConf $\uparrow$ & AVOffset($\rightarrow 0$) & M-LMD $\downarrow$ & F-LMD $\downarrow$ \\
    \midrule
    I & 7.359 & 7.586 & 0.000 & 0.000 & 0.000 \\
    II & 12.534 & 6.562 & 7.000 & 10.468 & 7.892 \\
    III & 9.300 & 5.344 & 3.000 & 4.532 & 5.246 \\
    IV & \textbf{8.145} & \textbf{6.288} & \textbf{0.000} & \textbf{3.354} & \textbf{3.465} \\
    \hdashline 
    V & \textbf{8.484} & \textbf{6.894} & \textbf{0.000} & \textbf{3.301} & \textbf{3.566} \\
  \bottomrule
\end{tabular}
}
\end{table}

\begin{figure}[H]
  \centering
  \includegraphics[width=0.45\textwidth]{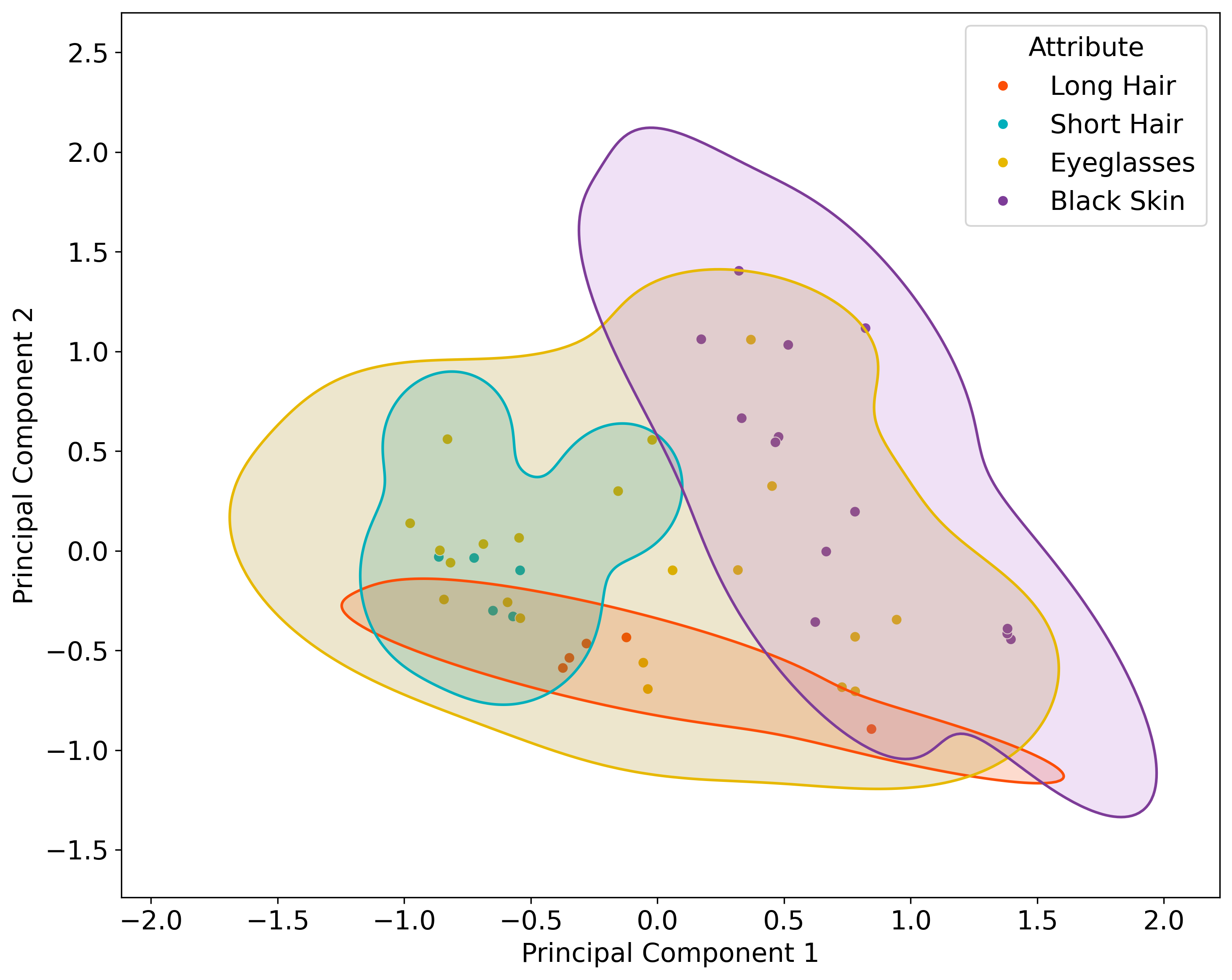}
  \caption{Principal Component Analysis (PCA) visualization of the four attributes.}
  \label{fig:pca}
\end{figure}

\begin{figure}[H]
  \centering
  \includegraphics[width=0.45\textwidth]{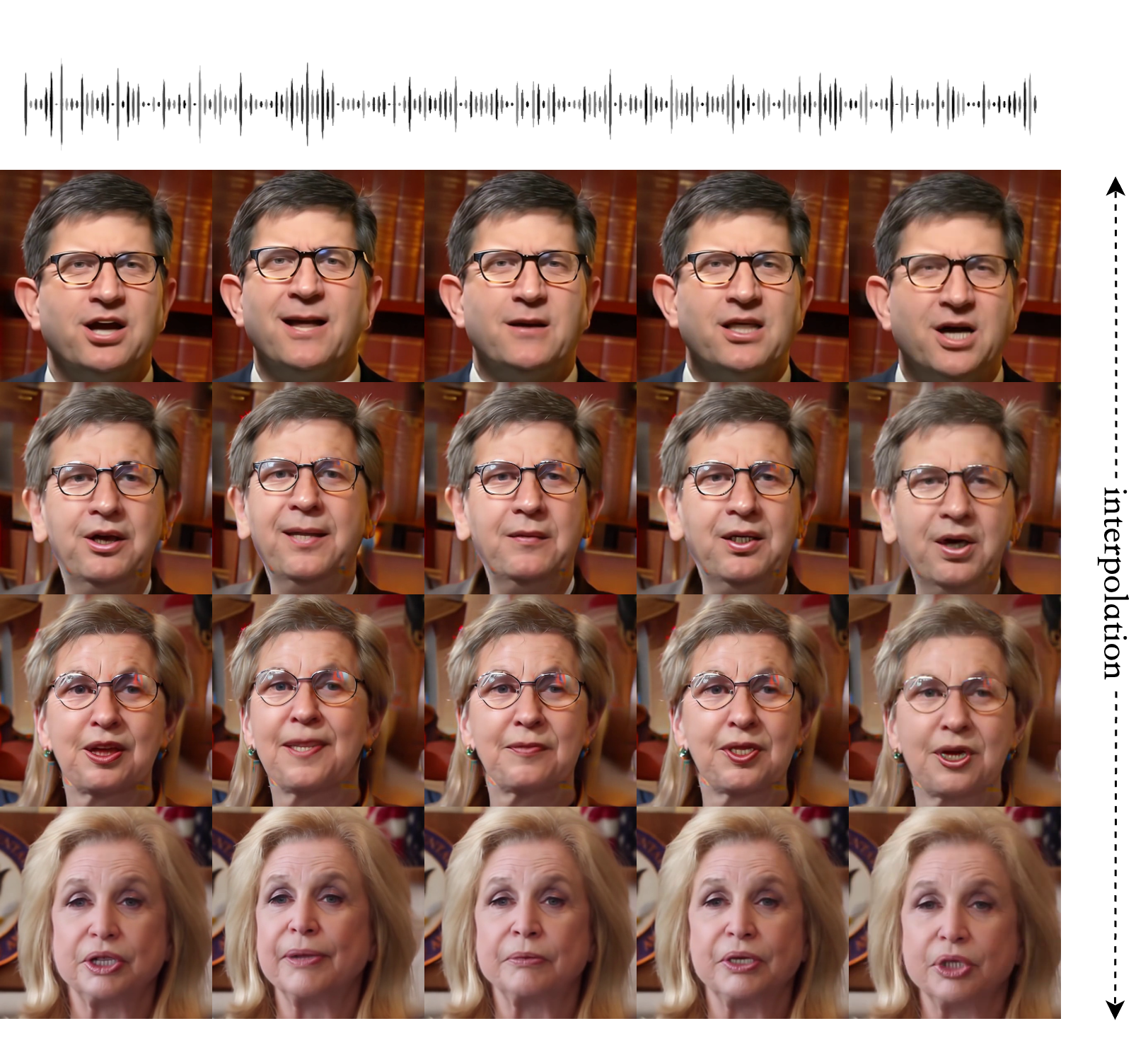}
  \caption{Talking head generation results after interpolating in the high-level semantic feature space using two reference images.}
  \label{fig:interpolation}
\end{figure}

\textbf{Linear Distribution of Attributes in Latent Space.} To verify the linear distribution of target attributes in latent space, we first visualized the latent space using Principal Component Analysis (PCA). Fig.\ref{fig:pca} shows that in the PCA space, samples of different attributes exhibit clear and meaningful separation along the principal component directions, proving that semantic latent variables exhibit a linear distribution in the latent space. This validates the rationality of linear operations in the semantic space and supports the classifier’s ability to distinguish between different attribute categories. Additionally, to demonstrate the interpolation capability of high-level semantic features across different attributes, we generated videos by interpolating features from two different speakers' images. The interpolation results appeared very natural in Fig.\ref{fig:interpolation} , showing good separation of attributes.

\section{Conclusion}
\label{sec:conslusion}
\subsection{Conclusion}
In summary, we have introduced a novel framework for editable talking face generation that significantly enhances both the realism and controllability of facial animations. By leveraging a combination of disentangled latent representations and fine-grained audio-visual alignment, our method enables intuitive editing capabilities such as face editing and lip-sync correction. Extensive experiments on representative baseline methods demonstrate that our approach not only achieves superior performance compared to existing representative methods, but also allows for diverse and personalized talking face generation. We believe this framework opens up promising directions for future research in personalized avatars, virtual assistants, and digital human synthesis.

\subsection{Limitation}
Despite the success of our framework, we acknowledge several limitations. First, although it demonstrates strong generalization ability, performance may degrade in scenarios involving highly diverse identities, complex head movements, or challenging conditions insufficiently represented in the training data. Second, the diffusion-based generation process, while producing high-quality results, incurs substantial computational costs, limiting its applicability in real-time settings. Third, our method currently supports only 40 predefined facial attributes and does not allow personalized image generation conditioned on arbitrary text prompts. Moreover, while directly extracting facial keypoints from audio improves lip-sync accuracy, it often leads to rigid or overly uniform facial expressions, reducing the realism of generated videos—as reflected in the user study results (Table~\ref{tab:user_study}). In the future, we aim to leverage models such as CLIP to enable flexible facial attribute editing through arbitrary text descriptions and to explore strategies that jointly preserve accurate lip synchronization and natural facial dynamics.

\subsection{Ethical Consideration}
We carefully consider the ethical implications of our work. Methods such as FaceEditTalker, which enable facial attribute editing, entail inherent risks, including unauthorized use of personal likeness and the spread of misleading content. To address these concerns, we restrict our framework strictly to academic research and prohibit any fraudulent or deceptive applications. The trained model and supporting detection resources will be shared only with the deepfake detection research community to strengthen identification efforts and promote the responsible advancement of this technology.


\section{Acknowledgments}
The work was jointly supported by the National Natural Science Foundations of China under grants No. 62272364, 62472342, the Provincial Key Research and Development Program of Shaanxi under grant No. 2024GHZDXM-47, the Research Project on Higher Education Teaching Reform of Shaanxi Province under Grant No. 23JG003.

\bibliographystyle{IEEEtran}
\bibliography{references}

\vspace{-15pt}

\begin{IEEEbiography}[{\includegraphics[width=1in,height=1.25in,clip,keepaspectratio]{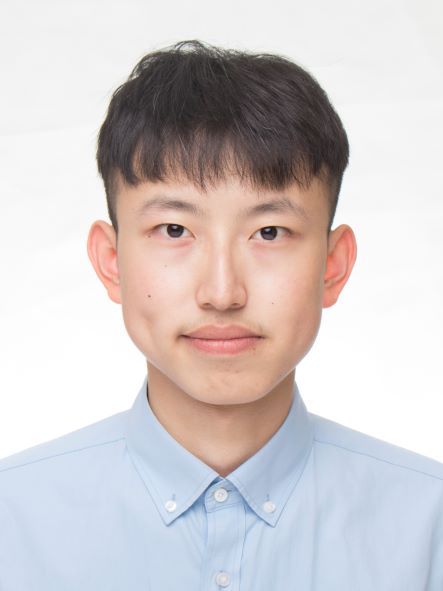}}]{Guanwen Feng}
received the B.S. degree in software engineering from Hangzhou Dianzi University in 2020. He is currently pursuing the Ph.D. degree with the School of Computer Science and Technology, Xidian University. His research interests include talking face animation, sign language generation, and traffic prediction.
\end{IEEEbiography}

\vspace{-15pt}

\begin{IEEEbiography}[{\includegraphics[width=1in,height=1.25in,clip,keepaspectratio]{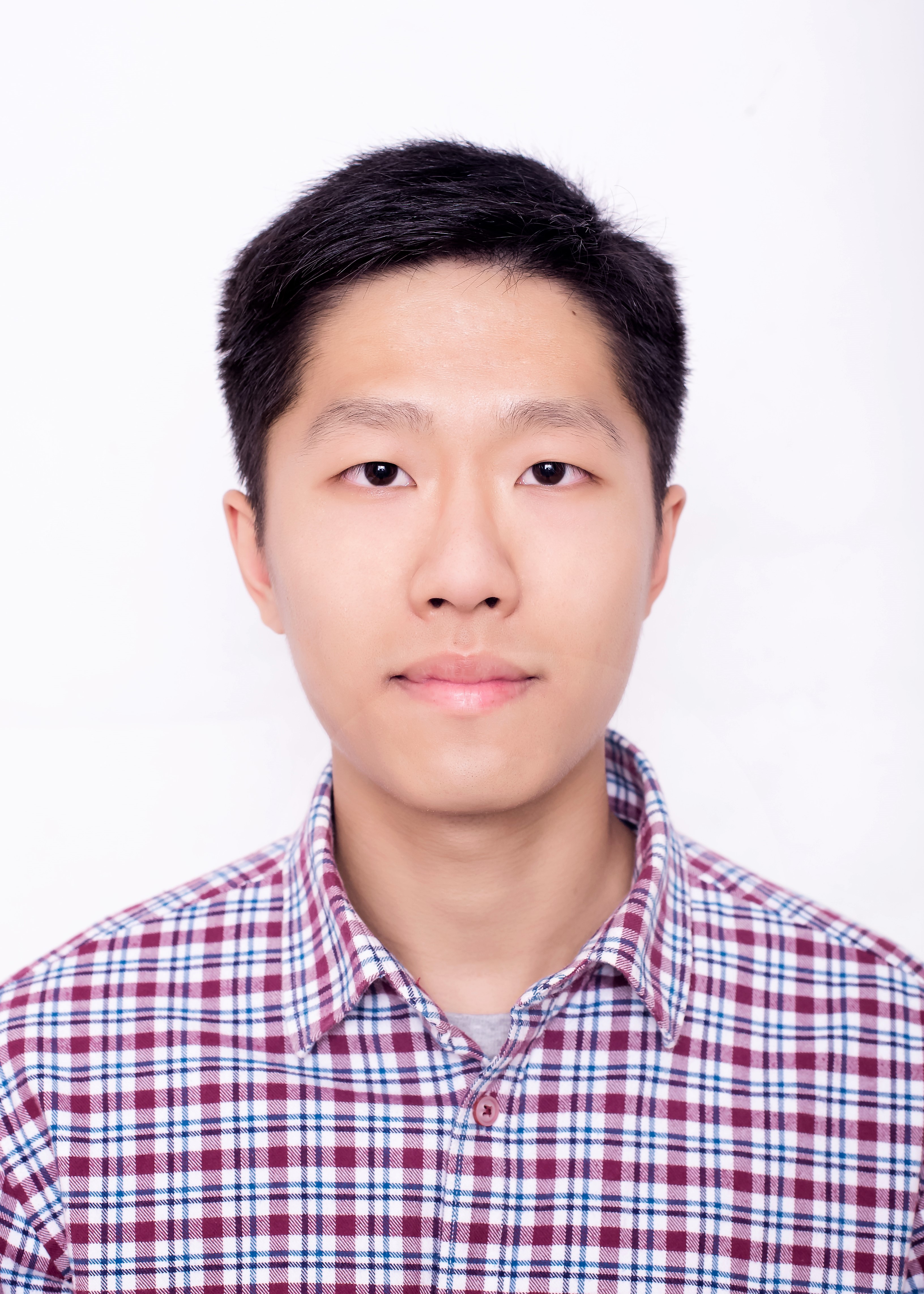}}]{Zhiyuan Ma}
received the B.S. degree in computer science and technology from Zhejiang Gongshang University in 2021. He is currently pursuing the M.S. degree at the School of Computer Science and Technology, Xidian University. His research interests include talking head generation and object detection.
\end{IEEEbiography}

\vspace{-15pt}

\begin{IEEEbiography}[{\includegraphics[width=1in,height=1.25in,clip,keepaspectratio]{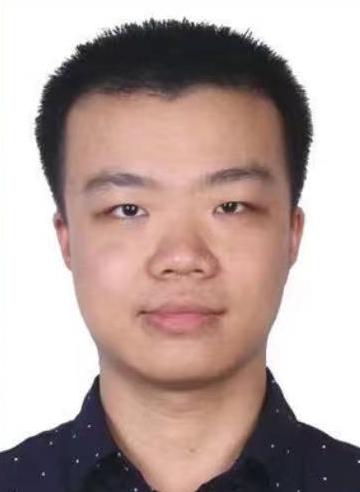}}]{Yunan Li}
received B.S. and Ph.D. degrees from the School of Computer Science and Technology, Xidian University, Xi’an, China, in 2014 and 2019, respectively. He is currently a Huashan Elite Associate Professor at Xidian University. His research interests include computer vision and pattern recognition, with a focus on applications in image enhancement and action/gesture recognition.
\end{IEEEbiography}

\vspace{-15pt}

\begin{IEEEbiography}[{\includegraphics[width=1in,height=1.25in,clip,keepaspectratio]{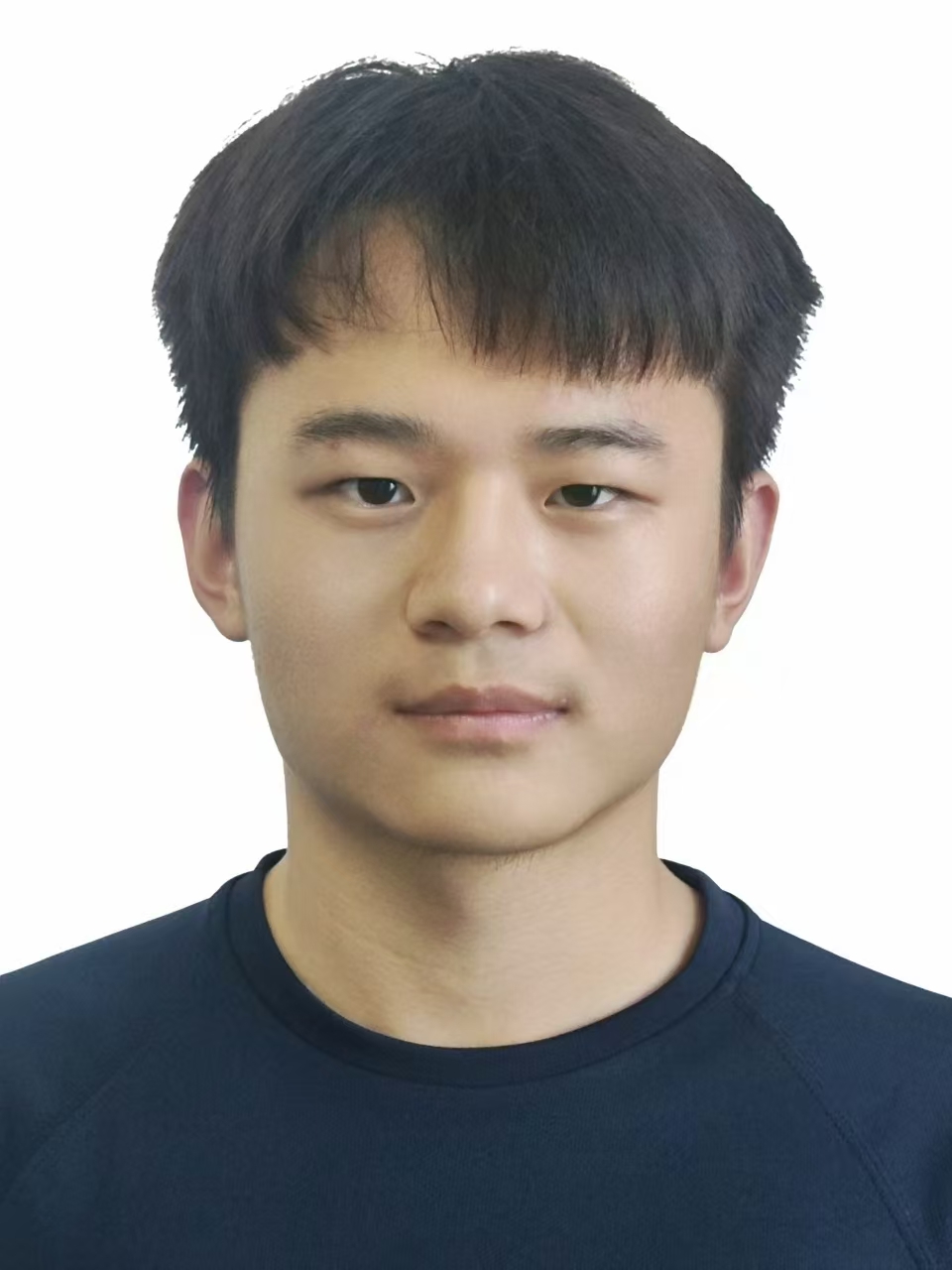}}]{Jiahao Yang} 
is a third-year undergraduate majoring in Big Data Management and Applications at the School of Economics and Management, Xidian University. His research interests include talking head generation.
\end{IEEEbiography}

\vspace{-15pt}

\begin{IEEEbiography}[{\includegraphics[width=1in,height=1.25in,clip,keepaspectratio]{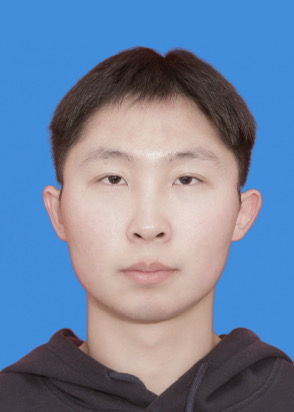}}]{Junwei Jing} 
is currently an undergraduate student at the School of Computer Science and Technology, Xidian University. His research interests include sign language generation and multimodal learning.
\end{IEEEbiography}

\vspace{-15pt}

\begin{IEEEbiography}[{\includegraphics[width=1in,height=1.25in,clip,keepaspectratio]{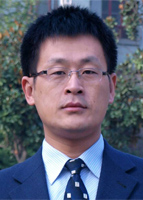}}]{Qiguang Miao}
is a professor and Ph.D. supervisor with the School of Computer Science and Technology, Xidian University. He received his Ph.D. degree from Xidian University in 2005. His research interests include intelligent image/video understanding and big data. In recent years, he has published more than 100 papers in leading international journals and conferences.
\end{IEEEbiography}
\vfill

\end{document}